\documentclass{article} 
\usepackage{nips13submit_e,times}
\usepackage{hyperref}
\usepackage{url}
\usepackage{amsfonts}
\usepackage{amsmath}
\usepackage{algorithm}
\usepackage{algorithmicx}
\usepackage{algpseudocode}
\usepackage{graphicx}
\usepackage{pdflscape}

\usepackage{booktabs}
\usepackage{colortbl}
\usepackage{color}
\def\arraystretch{1.2}

\title{Multi-Objective Evolutionary Beer Optimisation}

\author{
Mohammad Majid al-Rifaie\thanks{Corresponding author.} \hspace{0.3cm}
Marc Cavazza \vspace{5mm} 
\\ 
School of Computing \& Mathematical Sciences, University of Greenwich, London, UK
\\
\small{\texttt{ M.AlRifaie@gre.ac.uk }}\\
National Institute of Informatics, Tokyo, Japan\\
\small{\texttt{ cavazzam@acm.org }}
}

\nipsfinalcopy 

\begin{document}
\maketitle

\begin{abstract}
Food production is a complex process which can benefit from many optimisation approaches. However, there is growing interest in methods that support customisation of food properties to satisfy individual consumer preferences. This paper addresses the personalisation of beer properties. Having identified components of the production process for beers, we introduce a system which enables brewers to map the desired beer properties into ingredients dosage and combination. Previously explored approaches include direct use of structural equations as well as global machine learning methods. We introduce a framework which uses an evolutionary method supporting multi-objective optimisation. This work identifies problem-dependent objectives, their associations, and proposes a workflow to automate the discovery of multiple novel recipes based on user-defined criteria. The quality of the solutions generated by the multi-objective optimiser is compared against solutions from multiple runs of the method, and those of a single objective evolutionary technique. This comparison provides a road-map allowing the users to choose among more varied options or to fine-tune one of the preferred solutions. The experiments presented here demonstrate the usability of the framework as well as the transparency of its criteria.
\end{abstract}

\section{Introduction}
\label{sec:introduction}
Optimisation of food production processes, given its real-world significance, has been seen primarily through the lens of process optimisation with the sole objective of finding solutions that meet the precise characteristics of the product. 
Acknowledging the potential for several viable solutions when optimising the food processes, this real-world problem poses itself as a challenging tasks with an inherently underdetermined characteristic~\cite{donoho2012sparse,phillips2014best}. This work addresses the issue through a workflow based on a multi-objective optimiser, non-dominated sorting genetic algorithm II or NSGA-II to (1) generate solutions with the required characteristics defined by the domain experts, and (2) investigate solutions quality. 

We use brewing as an application domain, as it follows a well-defined process yet allows many variants of the end-product to be generated based on a range of ingredients and parameters of the brewing process itself. The common practice of the strict reliance on the existing beer recipes (as opposed to exploring novel ones) was dictated by the complexity of the process which is unforgiving towards errors~\cite{steenackers2015chemical}. This is particularly the case when the principal goal is the production of a product with precise and stable physio-chemical properties.

In order to progress from current practice, the aim of this work is to propose a brewing optimisation system that returns several realistically acceptable ingredient combinations when these exist. This is primed by the finding that brewing allows several levels of experimentation, which are however not directly compatible with a scalable production process. Our contribution would thus consists in reconciling optimisation and exploration of new recipes and recipe space.

The proposed method aims to map the brewing elements onto target properties. Basic knowledge is acquired through `reverse engineering' of some well-recognised labels based on their known characteristics. These characteristics allow both the validation of the generated solutions, and lay the foundations to reformulate the task as a multi-objective problem. The motivation is to demonstrate the system's ability in offering various solutions, in order to allow the domain experts the ease and flexibility of choosing from a suite of validated solutions; this is of practical relevance when either the diversity of the solutions is a priority, or the ability to fine-tune one or more of the identified solutions.

In this paper, previous related research are presented in Section~\ref{sec:background} which are then followed by a summary of some of the key concepts and formulas guiding the fermentation process. This enables an introduction to the objectives associated to the optimisation process. Subsequently, the evolutionary methods used to address the problem are presented. The experiments and results of `cloning' twenty products are reported in Section~\ref{sec:exp_n_res} where search and objective spaces are investigated. Directions for extending our results are suggested in the conclusion.

\section{Background}
\label{sec:background}
Several attempts have been made at optimising various elements of beer brewing; this is because of the its mix of standardisation and ingredient-based variability. Previous work have however, most often, considered specific or causal relationships between ingredients and individual properties known to play a significant role in consumers' preferences (e.g. foamability, flavour profile, temperature, aroma) or process parameters (e.g. processing time). These work can be categorised as process-centric, when the objective is process improvement, often with the intention to partially automate, or outcome-centric when researching the determinants of beer quality and organoleptic properties. In the latter case, focus can be on specific determinants or adopt a more global perspective on the end-product.

Beer {foamability} is studied in~\cite{viejo2018robotics} where robotics and computer vision techniques are combined with non-invasive consumer biometrics to assess quality traits from beer foamability. It is known that foam-related parameters are associated with beer quality and dependent on the protein content. A recent study explored the development of a machine learning model to predict the pattern and presence of 54 proteins~\cite{gonzalez2020development}. 
Furthermore, in another study, an objective predictive model is developed to investigate the intensity levels of {sensory descriptors} in beer using the physical measurements of colour and foam-related parameters, where a robotic pourer was used to obtain some colour and foam-related parameters from a number of different commercial beer samples~\cite{gonzalez2018assessment}. It is also claimed that this method could be useful as a rapid screening procedure to evaluate beer quality at the end of the production line. 

Another recent work centred on process optimisation~\cite{gonzalez2020digital} investigated sonication and traditional brewing techniques, with results demonstrating that the models developed using supervised machine learning based on near-infrared spectroscopy can accurately estimate physico-chemical parameters.

On a related topic, {mash separation} is known to be a critical pre-processing step in beer production where a high-quality stream of solubilised grain carbohydrates and nutrients is fed to the fermentors. Recent work by Shen et al.~\cite{shen2019statistical} performed a sensitivity analysis towards mash separation improvements. It concluded that strong wort volume and incoming feed quality to the mash filter have the strongest effect on filtration time, which it sees as a key performance metric for process optimisation. Some recent research have also been exploring faults detection in beers using artificial intelligence methods, as well as using strain development methodology to breed industrial brewing yeast~\cite{gonzalez2021smart, krogerus2021efficient}.

In another beer optimisation related task by Charry-Parra et al.~\cite{charry2011beer}, a technique was developed for the identification and quantification of volatile compounds of beer. 
This validation methodology enables its use as a quality control procedure for beer {flavour} analysis. 
A computational implementation of a kinetic model has also been proposed to rapidly generate temperature manipulations, simulating  the  operation  of  each  candidate profile~\cite{rodman2016multi}.

Compared to process improvement, optimising the {aroma} profile has been dedicated less work. For instance, Trelea et al.~\cite{trelea2004dynamic} obtained various desired final aroma profiles while reducing the total processing time using dynamic optimisation of three control variables: temperature,  top  pressure  and  initial  yeast  concentration  in  the  fermentation  tank; the  optimisation  is  based  on  a  sequential  quadratic  programming  algorithm  on  top of a  dynamic  model  of  alcoholic  fermentation  and  on  an  aroma  production model. Another recent work assesses the final aromatic profiles and physicochemical characteristics of beers~\cite{viejo2020beer}. This work presents artificial intelligence models based on aroma profiles, chemometrics, and chemical fingerprinting obtained from 20 commercial beers used as targets. 

Ermi et al.~\cite{ermi2018deep} explored two deep learning architectures  with the aim of  \textit{classifying}  coarse- and  fine-grained  beer types  and  \textit{predicting  ranges}  for  original  gravity~(OG),  final  gravity~(FG), alcohol by volume (ABV), international bitterness units (IBU), and colour. 

In summary, previous work have used various optimisation techniques, sometimes combined with modelling approaches (structural equations), targeting fermentation, foam, aroma profile, predicting beer flavours, , and controlling of beer fermentation process. 

The majority of approaches can still be categorised as traditional optimisation models, some based on process (e.g. kinetic) modelling, and others on specific relationship between limited number of key variables. To some extent, even previous use of DL was more focused on establishing direct relationships between parameters than uncovering global behaviour. They remain relevant in their identification of key elements of the process, parameters and local optimisation issues. However, there is a need for a more global approach, unifying local determinants within an outcome-centred optimisation approach. 

To address the novel challenge of reverse brewing, our work proposes to use a gradient-free, evolutionary approach to facilitate the discovery of validated, yet high fidelity and novel recipes, while taking into account both users preferences, and their constraints. Our approach takes advantage of the evolutionary methods in terms of optimisation dynamics and exploration of the solution space; this is not directly covered by the previous methods which have been mainly focusing on optimisation of specific aspects of the brewing processes, often starting from ingredients or recipe towards determining beer properties, whereas in this work, we take the reverse process of discovering recipes and the prerequisite ingredients from user tailored beer properties.

\subsection{Process Equations and Objectives} \label{sec:formulas}

Before articulating the technical assumptions underpinning the optimisation framework, we need to introduce some relevant elements of the underlying application. This will also constitute a first description of the relevant parameters and their dependencies.

The key objectives contributing towards the optimisation process are original gravity~(OG) which is the gravity of the wort pre-fermentation, final gravity~(FG) referring to the gravity post fermentation, alcohol by volume~(ABV), bitterness or international bitterness unit~(IBU) and colour which are used by the optimiser to determine the suitability of each proposed solution.
In the brewing process, ingredients are divided in three broad categories: hops, fermentables and yeasts. In addition to weight, several other relevant features are needed to calculate their impact on the brewing process (e.g. hops: alpha and beta and time values; fermentable: yield, colour, moisture and diastatic power; yeast: minimum and maximum temperatures, and attenuation). Beer's taste changes significantly depending on the exact quantities and varieties of ingredients and their timing in the process. 
From a food science perspective, the brewing process, although in some parts empirical, has been the subject of many descriptions and partial formalisation which are however sufficient to derive relevant equations. The overall process can thus be described as a set of elementary transformations that can be integrated within an optimisation framework. More specifically, a number of formal relationships between ingredients and target properties are sufficiently specific to support the generation of fitness functions. Some of the relevant formulas are discussed next.

{FG} $=f(\text{OG},\text{Yeast Attenuation})$, the gravity (i.e. sugar leftover from malt) after the fermentation, is defined as:
\begin{eqnarray}
\text{FG} = \text{OG} - (\text{OG}-1) \times \text{Yeast Attenuation} \label{eq:FG}
\end{eqnarray}

Gravity refers to the density of the solution and two items impact the overall density of wort: water, with density of $1.000$, and sugar from the malt. Given the heavier and larger molecule of sugar, when dissolved in water, the overall density increases. However, a third item, ethanol, which is produced by the yeast, has a lighter density of $0.789$. Subsequently, during the fermentation process, when the sugar is consumed, ethanol is produced, which in turn reduces the overall density of the solution, leading to $\text{FG} < \text{OG}$~\cite{palmer2017brew}. A dry or crisp flavor is the result of a lower FG, while a sweet or malty flavor has a higher FG. The distance between OG and FG determines the amount of alcohol in the product.

{ABV} $=f(\text{OG},\text{FG})$ and is defined as~\cite{hall1995brew,daniels1998designing}: 
\begin{eqnarray}
\text{ABV} =\frac{76.08~(\text{OG}-\text{FG})\text{FG}} {0.794~(1.775-\text{OG})}\label{eq:abv2}
\end{eqnarray}

{IBU}
is determined by taking into account the bitterness produced by hops or the hop extracts (from the fermentables), thus $\text{IBU} = f(\vec{\text{hops}},\vec{\text{fermentables}},\text{volume})$. The bitterness produced by hop is calculated as follows: 
\begin{eqnarray}
\text{IBU}_h = \sum_{i = 1}^{N_h} \frac{10w_i\alpha_i(1 - \exp^{-0.04 t_i})}{4.15~v} 1.65 \times 0.000125^{(\text{OG}-1)} 
\end{eqnarray}
where $N_h$ is the number of hops; $\vec{w}$ represents the weight; $v$ is the volume or batch size; $\vec{t}$ is time in minutes; and fermentables' bitterness is defined as:
\begin{eqnarray}
\text{IBU}_{f} = \sum_{i = 1}^{N_f} \frac{g_i~w_i}{v}
\end{eqnarray}
\noindent where $N_f$ is the number of fermentables; and $\vec{g}$ is `IBU gal per lb' which is associated with each fermentable and is known for each ingredients.
The final IBU is the sum of the individual IBUs: $\text{IBU} = \text{IBU}_h + \text{IBU}_f$.

\noindent{IBU/GU}
is often described in the following categories: cloying, slightly malty, balanced, slightly hoppy, extra hoppy, and very hoppy. $\text{IBU/GU} = f(\text{OG},\text{IBU})$:
\begin{eqnarray}
\text{IBU/GU} = \frac{\text{IBU}}{1000(\text{OG} - 1)} 
\end{eqnarray}
{Colour} is mainly determined by malts and hops.
The two main protocols to measure colour are Standard Reference Method~(SRM) and European Brewing Convention~(EBC). 
SRM, which is used in this work, was initially adopted in 1950 by the American Society of Brewing Chemists. 
The value of SRM is determined by measuring the attenuation of light of a particular wavelength ($430$ nm) in passing through $1$ cm of the beer, expressing the attenuation as an absorption and scaling the absorption by a constant ($12.7$ for SRM or $25$ for EBC, where $\text{EBC} = \text{SRM} \times 1.97$).
Stone and Miller~\cite{stone1949standardization} proposed malt colour unit~(MCU), which is defined as:
\begin{eqnarray} \label{eq:MCU}
\text{MCU} = \sum_{i=1}^{N_f} \frac{c_i~ w_i}{v}
\end{eqnarray}
\noindent where $\vec{c}$ refers to grains' colour (fermentables' colour). 
As shown in the equation above, for more than one grain type, the MUC is calculated for each and all the values are aggregated.
However, MUC tends to overestimate the colour value for darker beers (MUC $>10.5$). 
Thus, Morey~\cite{morey2000hop} derived the following to deal with SRM up to $50$: 
\begin{eqnarray}\label{eq:COLOUR}
\text{SRM} = 1.4922 \sum_{i=1}^{N_f} \frac{c_i~ w_i^{0.6859}}{v}
\end{eqnarray}

\subsection{Evolutionary Optimisation}

We propose an optimisation framework which in essence unifies the various elements of the brewing process from the individual reaction mechanisms to the articulation between the various parameters. This framework relies mainly on the multi-objective optimiser, yet provides both a comparison and a potential integrative workflow with a single objective evolutionary optimiser. These algorithms are presented next.

\subsubsection{NSGA-II}

 Non-dominated sorting genetic algorithm II~\cite{deb2002fast} is an efficient methods dealing with multi-objective optimisation problems. 
 This method identifies non-dominated solutions which represent different trade-offs between a number of objectives, the optimal trade-off being represented by a set of Pareto solutions generated. In addition to using the standard genetic algorithm processes (mutation, crossover and selection), it benefits from an elitist strategy, utilising diversity preserving mechanism, and deploying non-dominated solutions, where individuals are selected based on their crowding distances and non-dominated ranks. Moreover, previous work has demonstrated the potential for NGSA-II to work alongside other analytic or decision-making tools.

\subsubsection{DE}

The Differential evolution algorithm is used to provide a comparative analysis of the results\footnote{For a detailed study of the experiments limited to single objective optimisers, see~\cite{alRifaie_2020_beer_GECCO,alRifaie2022_foods}. This analysis shows the superiority of one of the single objective methods introduced in~\cite{alRifaie_2014_DFO} with its exploration-exploitation study investigated in~\cite{alRifaie_2021_Entropy_DFO_zones,alRifaie_2021_CEC_uDFO}. } in some aspects of the experiments. In this work, from the different varieties of the algorithm, DE/best/1 version, known for its competitiveness and robustness, is used~\cite{das2010differential}. The cross-over rate and the differential weight are $C_R \in [0,1]$ and $F \in [0,2]$.

\section{Experiments and Results} 
\label{sec:exp_n_res}

This section presents a set of experiments where physico-chemical properties of several commercial beers are used along with the in-stock inventory to evaluate the proposed system by ``reverse manufacturing'' these commercial beers from their target physico-chemical properties.
The list of ingredients in this experiment is shown in Table~\ref{tab:stockII}, and the desired physio-chemical properties are selected from the list of twenty commercial beers as shown in Table~\ref{tab:RealBeerFeatures}.

In this work, the optimiser takes an inventory of the existing ingredients and their weights (as dimensionality of the problem and bounds to each dimension respectively) along with a desired set of organoleptic properties for a particular product (as the means to measure solution fitness), and returns an optimal set of ingredient lists and their associated amounts (as solution vectors) which facilitate the production of the target product.

The experiments conducted in this section are to cater for a proof of principle study, which simulate a realistic scenario in a small-scale brewery, where the brewer's efficiency is set to $58\%$\footnote{This refers to the efficiency of equipment in extracting sugars from malts during the mashing stage. Efficiency is higher for larger-scale industrial setups.}, boil size of $24 L$, batch size of $20 L$, and boil time of $60$ minutes. Despite following a simulative approach, the grounding for our experiments is found in the pre-existing descriptions associated to commercial products being ``reverse engineered''. This covers all the relevant parameters as well as user preferences and organoleptic properties through respectively published ingredients, label contents, advertising material and product reviews. 

\subsection{Experiment Setup}

The experiments reported here give an indication of the overall performance of the system on each product when generating solutions whose diversity and quality will be analysed. Based on the experiment results and depending on domain experts' needs, recommendation on brewing strategies will be provided. These strategies depend on whether users are interested in exploring a varied set of solutions, or to fine-tune a preferred solution from the solution space.

The population size for NSGA-II is set to $100$, generating the same number of offsprings. In each generation, duplicate individuals are removed, and given the continuous nature of the solution space, simulated binary crossover and polynomial mutation 
are used~\cite{deb2007self}. The crossover probability is set to $p_c=0.9$ 
and using distribution indexes~\cite{deb1995simulated}, the crossover and mutation operators are set to $\eta_c=15$ and $\eta_m=20$ respectively. The termination criterion is set to $1000$ generations. The population size in DE is $100$, and $F$ and $C_R$ are set to 0.5. The termination criteria are set to either reaching 100,000 function evaluations or the error threshold defined below. 

In the experiments, at the end of each run, the number of successful solution vectors in the population with the overall error of $e\le 0.05$ (which is defined next) is used as one of the performance measures, along with the count of non-dominated solutions for NSGA-II. 
There are 30 independent runs for each experiment and the results are summarised over these independent simulations. 

In this work, given OG, FG and ABV are interdependent (see Eq.~\ref{eq:abv2}), the task is formulated into a 3-objective problem with $f_{1}$, which takes into account the Euclidean distance between the desired OG, FG, ABV and the corresponding generated values; $f_{2}$, the proximity to user-specified IBU, and $f_{3}$ which returns the distance from the desired colour.
\begin{align*}
f_1(\vec{x}) &= \sqrt{ (f_{\text{OG}}(\vec{x}) - \text{OG})^2 + (f_{\text{FG}}(\vec{x}) - \text{FG})^2 + (f_{\text{ABV}}(\vec{x}) - \text{ABV})^2 }\\
f_2(\vec{x}) &= \sqrt{ (f_{\text{IBU}}(\vec{x}) - \text{IBU})^2 }\\
f_3(\vec{x}) &= \sqrt{ (f_{\text{SRM}}(\vec{x}) - \text{SRM})^2 }
\end{align*}

\noindent where $\vec{x}$ is the list of ingredients, and OG, FG, ABV, IBU and SRM represent the desired values for a product as provided by the brewers (in this case from Table~\ref{tab:RealBeerFeatures}).

The overall error, $e=f_{1} + f_{2} + f_{3}$, is defined by the quality of the solution in terms of the proximity of the solution's fitness to the objective values. Note that $e$  is not used by the multi-objective optimiser to guide the optimisation.

\begin{table}
	\centering
	\caption{Input ingredients and their properties\label{tab:stockII}}
	\renewcommand{\arraystretch}{.8}
	\setlength{\tabcolsep}{3pt}
	\begin{tabular}{llrrcc}
		\toprule 
		\textbf{\#}  & \textbf{Hop}  & \textbf{Weight}  & $\alpha$  & $\beta$  & \textbf{Time }\tabularnewline
		\midrule 
		1  & Cascade  & 100 g  & 6  & 6  & 60\tabularnewline
		2  & Chinook  & 100 g  & 13  & 3.5  & 60\tabularnewline
		3  & Northern Brewer  & 100 g  & 9  & 4  & 60\tabularnewline
		4  & Magnum  & 40 g  & 13.5  & 6  & 60\tabularnewline
		5  & Fuggles  & 50 g  & 4.5  & 2.5  & 60\tabularnewline
		\bottomrule
		&  &  &  &  & \tabularnewline
		\toprule 
		& \textbf{Fermentable*}  & \textbf{Weight}  & \textbf{SRM}  & \textbf{Yield} & \textbf{Moisture}\tabularnewline
		\midrule 
		6  & Pale Malt (UK)  & 7 kg  & 3  & 78 & 0\tabularnewline
		7  & Caramel/Crystal Malt  & 1 kg  & 60  & 74 & 0\tabularnewline
		8  & Cara-Pils/Dextrine  & 1 kg  & 2  & 72 & 0\tabularnewline
		9  & Biscuit Malt  & 0.5 kg  & 23  & 79 & 0\tabularnewline
		10  & Wheat Malt (Belgium)  & 2 kg  & 2  & 81 & 0\tabularnewline
		11  & Chocolate Malt (UK)  & 0.5 kg  & 450  & 73 & 0\tabularnewline
		12  & Munich Malt  & 3 kg  & 9  & 80 & 0\tabularnewline
		13  & Pilsner (German)  & 5 kg  & 2  & 81 & 0\tabularnewline
		14  & Roasted Barley  & 0.5 kg  & 300  & 55 & 0\tabularnewline
		15  & Barley Flaked  & 0.5 kg  & 2  & 70 & 0\tabularnewline
		\bottomrule
		&  &  &  &  & \tabularnewline
		\toprule 
		& \textbf{Yeast}  & \textbf{Vol}  & \textbf{Min $^{\circ}$}  & \textbf{Max $^{\circ}$}  & \textbf{dB}\tabularnewline
		\midrule 
		16  & Safale S-04  & 11 mL  & 15$^{\circ}$  & 24$^{\circ}$  & 75\tabularnewline
		\bottomrule
	\end{tabular}
	
	\vspace{1mm}
	\raggedright{*~ IBU gal per lb is $0$ for all the fermentables.}
	
\end{table}

\begin{table}
	\centering
	\caption{Product list and their characteristics\label{tab:RealBeerFeatures}}
	\setlength{\tabcolsep}{4.4pt}
	\renewcommand{\arraystretch}{.8}
	\begin{tabular}{rlccccc}
		\toprule 
		\textbf{\#} & \textbf{Product Name} & \textbf{ABV} & \textbf{IBU} & \textbf{SRM} & \textbf{OG} & \textbf{FG}\tabularnewline
		\midrule 
		1 & Imperial Black IPA & 12.2 & 150 & 35 & 1.098 & 1.013\tabularnewline
		2 & Guinness Extra Stout & 5.1 & 40 & 40 & 1.070 & 1.034\tabularnewline
		3 & Atlantic IPA Ale & 8.4 & 80 & 13 & 1.074 & 1.013\tabularnewline
		4 & Tokyo Rising Sun & 15.4 & 85 & 71 & 1.125 & 1.023\tabularnewline
		5 & Punk Monk & 6.2 & 60 & 8.5 & 1.056 & 1.010\tabularnewline
		6 & Santa Paws & 4.7 & 35 & 22 & 1.048 & 1.013\tabularnewline
		7 & Sunmaid Stout & 11.1 & 50 & 100 & 1.102 & 1.026\tabularnewline
		8 & Vice Bier & 4.4 & 25 & 15 & 1.043 & 1.010\tabularnewline
		9 & Blitz Berliner Weisse & 4.3 & 8 & 4.5 & 1.040 & 1.007\tabularnewline
		10 & Jasmine IPA & 6.3 & 40 & 17.5 & 1.060 & 1.014\tabularnewline
		11 & No Label & 4.5 & 25 & 5 & 1.043 & 1.009\tabularnewline
		12 & Monk Hammer & 7.5 & 250 & 7.5 & 1.065 & 1.010\tabularnewline
		13 & Science IPA & 5.2 & 45 & 47 & 1.050 & 1.011\tabularnewline
		14 & Tropic Thunder & 7.5 & 25 & 86.36 & 1.074 & 1.020\tabularnewline
		15 & Blonde Export Stout & 7.7 & 55 & 8 & 1.075 & 1.020\tabularnewline
		16 & Indie Pale Ale & 4.8 & 30 & 8 & 1.044 & 1.008\tabularnewline
		17 & Funk X Punk & 7.2 & 42 & 12 & 1.058 & 1.004\tabularnewline
		18 & Atlantic IPA Ale & 8.4 & 80 & 28 & 1.074 & 1.013\tabularnewline
		19 & Kozel Dark   & 4.6 & 35.09 & 21.87 & 1.042 & 1.007\tabularnewline
		20 & Punk IPA & 5.6 & 40 & 7.6 & 1.053 & 1.011\tabularnewline
		\bottomrule 
	\end{tabular}
	
\end{table}
\renewcommand{\arraystretch}{.8}
\subsection{Results and Discussion}

Table~\ref{tab:successfulSol} reports the number of non-dominated individuals found at the end of the process; also, from these individuals, the ones with pre-determined proximity to the optimal values are indicated as successful solutions. During the optimisation process, the majority of individuals within the population approach constitute part of the Pareto front ($\ge 95\%$); however, not all return the desired overall error, $e$. As shown in the table, for the majority of the products, approximately $20-50\%$ of the individuals on the Pareto front are successful. In three of the products (i.e. 4, 7 and 14), no suitable solution is found. The reason behind these failures will be investigated by exploring each objective individually. 

\begin{table}
	\centering
	\caption{Summary of non-dominant and successful solutions\label{tab:successfulSol}}
	\setlength{\tabcolsep}{12pt}
	\begin{tabular}{c|cc|cc}
		\toprule 
		& \multicolumn{2}{c|}{\textbf{Non-dominant}} & \multicolumn{2}{c}{\textbf{Successful }}\tabularnewline
		\textbf{\#} & \textbf{Median} & \textbf{Stdev} & \textbf{Median} & \textbf{Stdev}\tabularnewline
		\midrule
		1 & 100 & 16.03 & 41.5 & 31.15\tabularnewline
		2 & 96 & 15.42 & 18 & 33.56\tabularnewline
		3 & 95 & 13.40 & 52.5 & 33.48\tabularnewline
		4 & 100 & 0 & 0 & 0\tabularnewline
		5 & 99 & 11.36 & 53.5 & 32.67\tabularnewline
		6 & 100 & 10.52 & 47 & 32.74\tabularnewline
		7 & 100 & 0 & 0 & 0\tabularnewline
		8 & 100 & 17.20 & 31 & 27.53\tabularnewline
		9 & 97 & 15.39 & 29 & 25.85\tabularnewline
		10 & 98 & 17.97 & 40 & 31.26\tabularnewline
		11 & 100 & 10.61 & 33.5 & 34.58\tabularnewline
		12 & 100 & 13.91 & 54.5 & 31.03\tabularnewline
		13 & 97.5 & 15.99 & 25.5 & 29.61\tabularnewline
		14 & 100 & 0 & 0 & 0\tabularnewline
		15 & 98.5 & 15.55 & 49.5 & 27.86\tabularnewline
		16 & 100 & 11.58 & 27.5 & 30.92\tabularnewline
		17 & 99.5 & 14.24 & 20.5 & 29.40\tabularnewline
		18 & 100 & 18.93 & 41.5 & 31.05\tabularnewline
		19 & 100 & 13.45 & 53 & 35.18\tabularnewline
		20 & 99.5 & 15.63 & 43.5 & 26.70\tabularnewline
		\bottomrule
	\end{tabular}
\end{table}

To evaluate the proximity of the objectives, $f_{1-3}$, of successful solutions in each run, initially the standard deviation of each objective is calculated in each run; these values are then averaged to measure the nearness of the objectives, or their deviations. 
Table~\ref{tab:obj-dis} reports the results, which illustrate the closeness of each independent objective values.

\begin{table}\centering
	\caption{Average objective deviation in each independent run
	\label{tab:obj-dis}}
	\setlength{\tabcolsep}{17pt}
	\begin{tabular}{cccc}
		\toprule 
		\textbf{\#} & \textbf{$f_{1}$} & \textbf{$f_{2}$} & \textbf{$f_{3}$}\tabularnewline
		\midrule 
		1 & 1.46E-04 & 3.88E-03 & 1.39E-03\tabularnewline
		2 & 1.34E-04 & 1.47E-03 & 1.90E-04\tabularnewline
		3 & 4.51E-04 & 6.09E-03 & 2.46E-03\tabularnewline
		4 &  NA  &  NA  &  NA\tabularnewline
		5 & 5.63E-04 & 7.09E-03 & 4.23E-03\tabularnewline
		6 & 6.74E-04 & 9.25E-03 & 4.10E-03\tabularnewline
		7 &  NA  &  NA  &  NA\tabularnewline
		8 & 7.86E-04 & 9.68E-03 & 5.20E-03\tabularnewline
		9 & 1.40E-03 & 7.05E-03 & 7.46E-03\tabularnewline
		10 & 7.62E-04 & 7.55E-03 & 3.77E-03\tabularnewline
		11 & 6.74E-04 & 5.90E-03 & 5.26E-03\tabularnewline
		12 & 2.50E-04 & 5.07E-03 & 1.07E-03\tabularnewline
		13 & 7.55E-04 & 9.25E-03 & 2.90E-03\tabularnewline
		14 &  NA  &  NA  &  NA\tabularnewline
		15 & 5.73E-04 & 8.48E-03 & 3.66E-03\tabularnewline
		16 & 1.20E-03 & 7.83E-03 & 6.39E-03\tabularnewline
		17 & 3.98E-04 & 4.42E-03 & 3.57E-03\tabularnewline
		18 & 3.94E-04 & 6.61E-03 & 2.51E-03\tabularnewline
		19 & 7.35E-04 & 5.86E-03 & 3.39E-03\tabularnewline
		20 & 8.06E-04 & 7.71E-03 & 4.71E-03\tabularnewline
		\bottomrule 
	\end{tabular}

\end{table}

In under-determined systems~\cite{donoho2012sparse} where there are fewer equations than unknowns (and by extension, more solutions to a problem), the proximity of the objectives in solution vectors does not strictly imply the proximity of solution vectors themselves.
To establish the distance of solution vectors within each run, a similar analysis is conducted, this time for each of the sixteen components in the solution vectors and the results are shown in Table~\ref{tab:sol-dis-x1-16}. It is evident that in each of the independent instances not only the distances between the objectives are modest, but also the solutions are not radically distant in the context of this real-world problem. This enables the optimiser to return several similar candidate solutions with delicate differences for the users to choose from depending on their production processes and criteria. This is a valuable feature which allows for a balanced choice to be made from the existing solutions.

\begin{figure*}
	\newlength{\imgW}
	\setlength{\imgW}{0.31\textwidth} 
	
	\begin{tabular}{ccc}
		\includegraphics[width=\imgW]{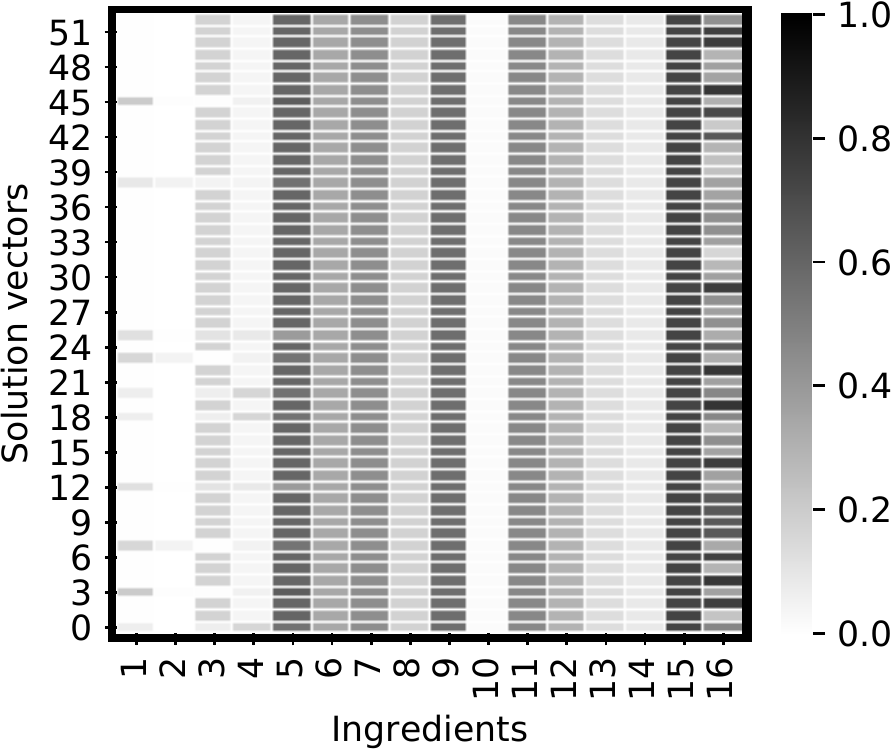}
		& \includegraphics[width=\imgW]{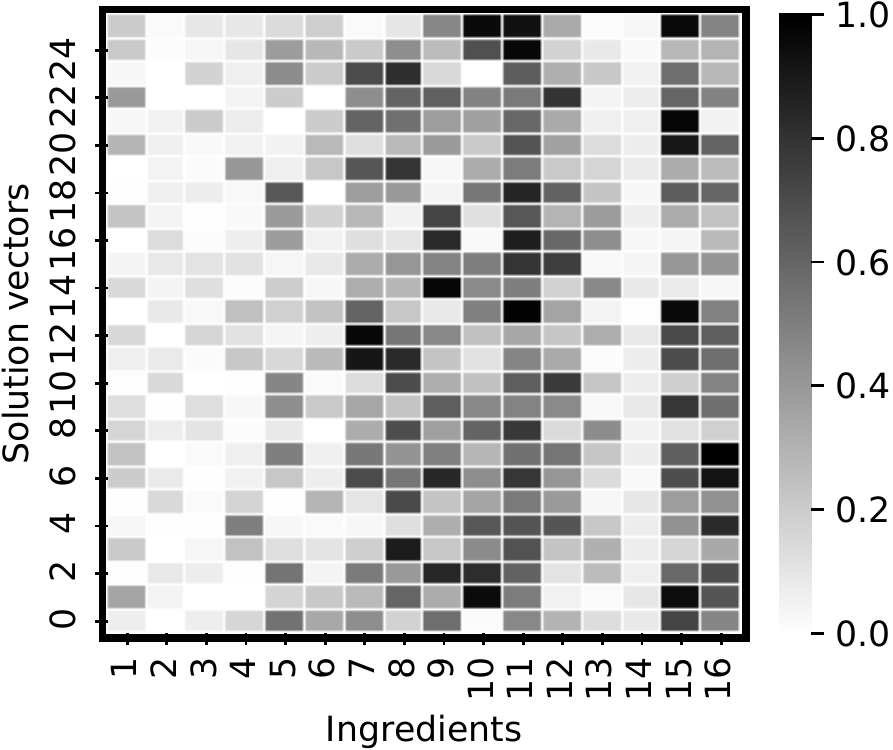} 
		& \includegraphics[width=\imgW]{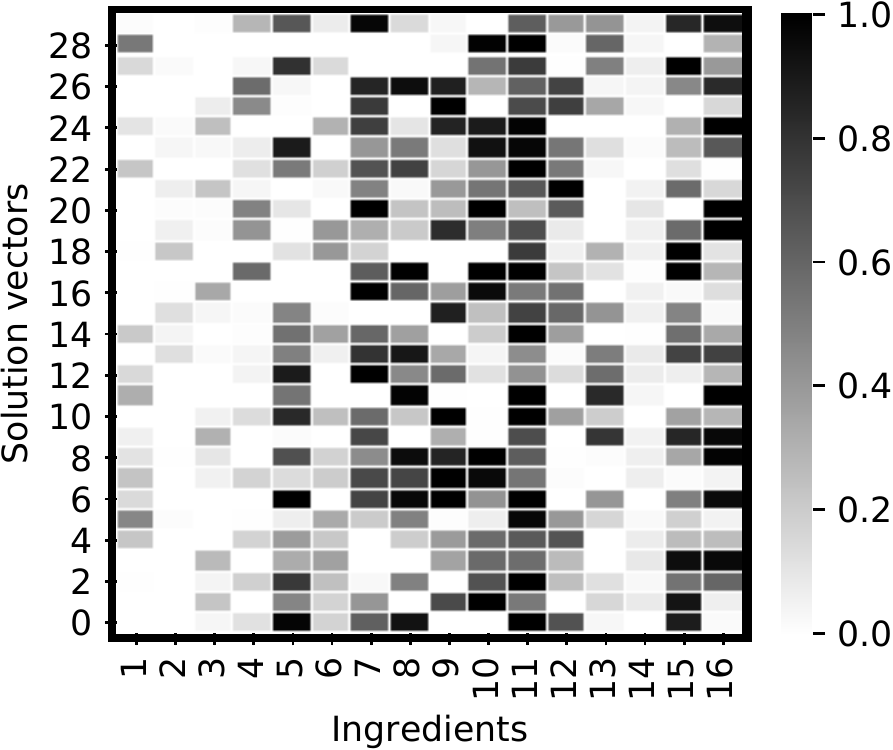} 
		\tabularnewline
		(a) & (b) & (c) \tabularnewline
	\end{tabular}

	\caption{Ingredients combinations, or solution vectors generated by (a) NSGA-II solutions in a single run returning 53 solutions (b)~NSGA-II solutions from multiple runs (30 independent runs) 25 of which return at least one successful solution to pick from, and (c) DE  solutions from multiple runs (30 independent runs) returning 30 successful solutions.  \label{fig:combinations_sols_NSGAII_SI}}
\end{figure*}

Fig.~\ref{fig:combinations_sols_NSGAII_SI}-a illustrates the closeness of solutions generated in a single NSGA-II run for one of the products, Guinness Extra Stout (the same observation is seen in the others products). This sample run, resulted in $53$ successful solutions; each line on the $y$-axis represents a solution vector which illustrates the amount of uptake in each ingredients on the $x$-axis; a darker shade indicates a higher uptake from the existing ingredient (e.g. ingredient 15 representing flaked barley) and a brighter shade highlights a lower uptake (e.g. ingredient 10 referring to wheat malt).

\begin{figure*}
	\setlength{\imgW}{0.31\textwidth} 

	\begin{tabular}{ccc}
		\includegraphics[width=\imgW]{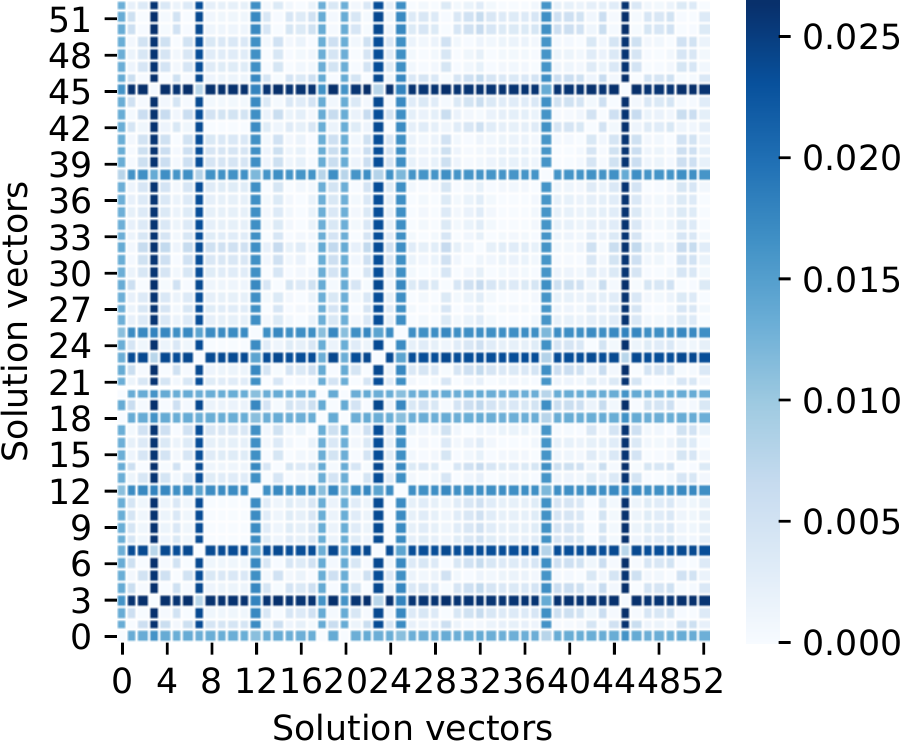}
		& \includegraphics[width=\imgW]{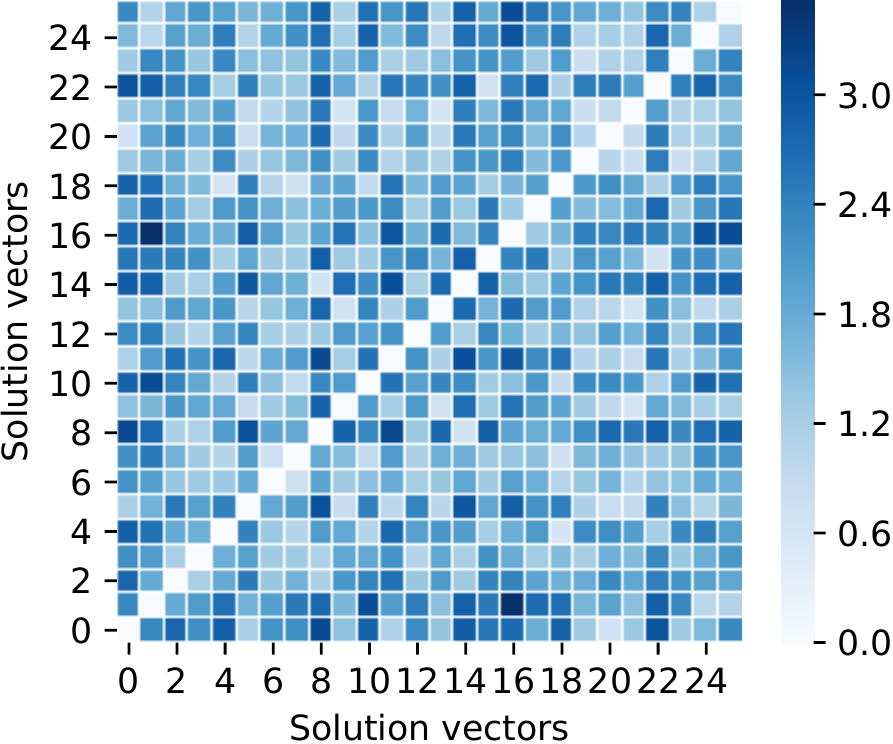} 
		& \includegraphics[width=\imgW]{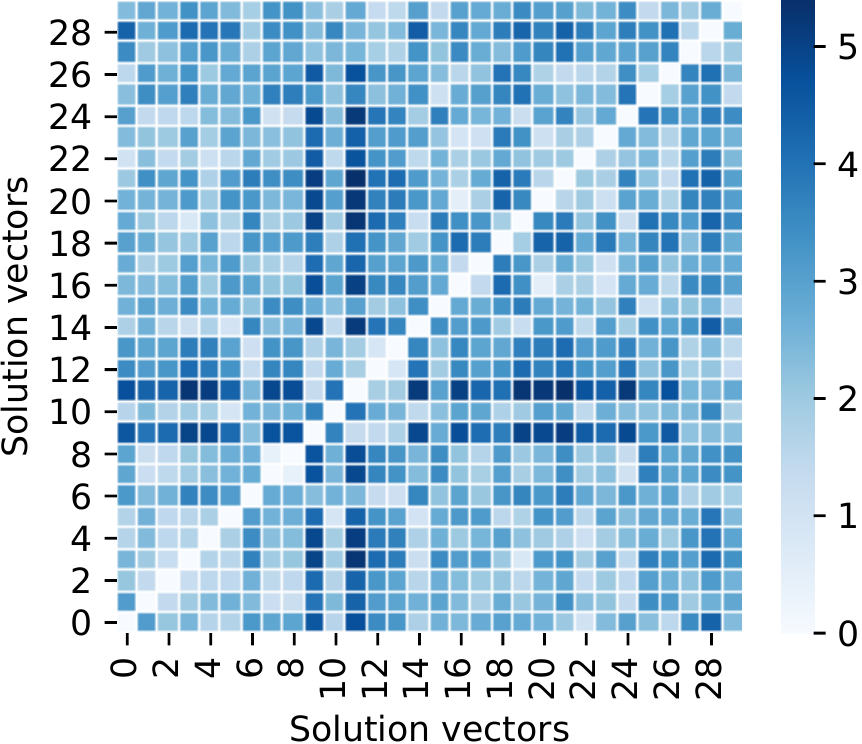} 
		\tabularnewline
		(a)  & (b)  & (c) \tabularnewline
	\end{tabular}
	
	\caption{Distance matrices for solutions generated by (a) NSGA-II solutions in a single run returning 53 solutions (b)~NSGA-II solutions from multiple runs (30 independent runs) 25 of which return at least one successful solution to pick from, and (c) DE solutions from multiple runs (30 independent runs) returning 30 successful solutions.  \label{fig:sols_diversity}}
\end{figure*}

To investigate whether more distinct solutions are available to this problem and if the optimiser (in each independent run) tends to `navigate' the individuals towards a particular `zone' within the Pareto front, the proximity of sampled solutions from each independent run is taken into account and visualised.
Taking a sample solution from each independent run (i.e. naturally, with varying starting points) demonstrates a greater distinctiveness in the diversity of the solution vectors (see Fig.~\ref{fig:combinations_sols_NSGAII_SI}-b). This confirms that NSGA-II, in the context of this problem, leads the population towards a particular segment of the Pareto front in each run. This experiment evidences the availability of more varied solutions in the solution space (utilising further crowd distancing measures and diversity preserving/promoting strategies in this context can be beneficial~\cite{coello2007evolutionary,tian2018strengthened,pamulapati2018i_}). To confirm this finding, DE, as the single objective evolutionary optimiser, is run 30 times and the generated solutions are visualised in Fig.~\ref{fig:combinations_sols_NSGAII_SI}-c.

\begin{figure}\centering
	\setlength{\imgW}{0.35\textwidth} 
	
	\includegraphics[width=\imgW]{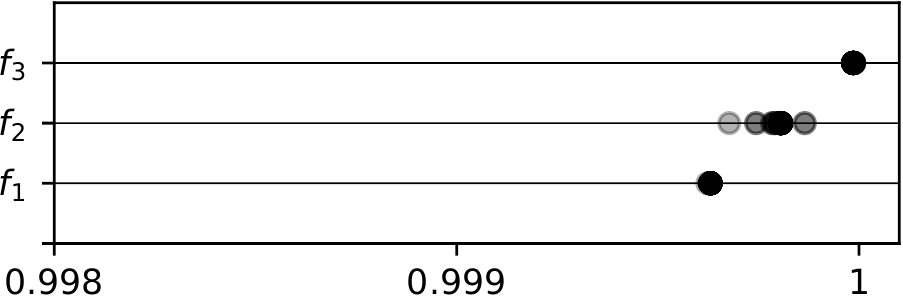}\\ \vspace{1mm}
	(a) Objectives in 53 solutions from in a single run 
	
	\vspace{3mm}
	\includegraphics[width=\imgW]{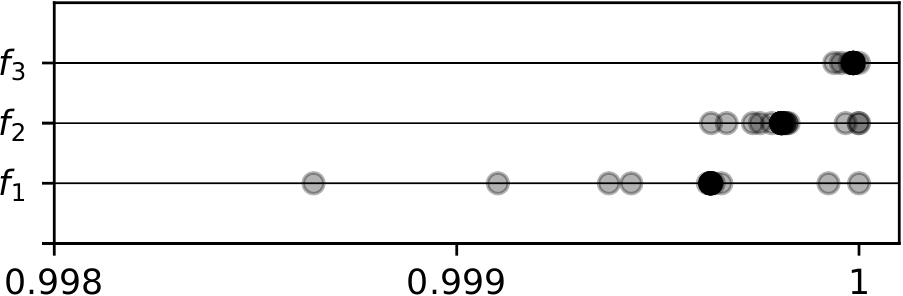}\\ \vspace{1mm}
	(b) Objectives in 25 solutions from 30 independent runs
	
	\caption{Distribution of objective values, $f_{1-3}$, in successful solutions generated by (a) single run of NSGA-II, and (b) multiple independent runs of NSGA-II. \label{fig:objs}}
\end{figure}

The observations on the performance of NSGA-II are extendable to the objective space, which as shown in Fig.~\ref{fig:objs}, demonstrate a stronger proximity in a single run NSGA-II, than sampling solutions from multiple runs of the algorithm.

To further evaluate the solutions' uniqueness, the distance between each pair is calculated. These values are presented as distance matrices and visualised in Fig.~\ref{fig:sols_diversity}, showing that depending of the user-specific needs, from less diverse to more diverse set of solutions, the following is recommended: single run of NSGA-II and choosing successful solutions; multiple runs of NSGA-II with a sample solution picked from each run; and multiple runs of single objective differential evolutionary optimiser. 
The maximum distances between solution vectors, which are shown in the labelled colour bar in Fig.~\ref{fig:sols_diversity}, are 0.0265 (Fig.~\ref{fig:sols_diversity}-a), 3.5156 (Fig.~\ref{fig:sols_diversity}-b) and 5.3905~(Fig.~\ref{fig:sols_diversity}-c) respectively.

In the results reported, there have been instances where the optimiser fails to suggest suitable ingredient combinations, more specifically, in three products. The average objective values in the three failed products in 30 runs is calculated, where each run results in 100 non-dominant individuals. Average objective distance of the individuals from the desired objective values is shown in Table~\ref{tab:obj-avg-failedProd}. 
Looking at the desired objective values in these products, there is a shared characteristic which prevents them from being optimised given the existing inventory. The reason lies in the high SRM values ($\ge 71$), which can only be achieved with greater availability of ingredients such as chocolate malt or roasted barley (i.e. ingredients 11 and 14 in Table~\ref{tab:stockII} respectively). The optimiser aims to increase the SRM value by using the other fermentables, resulting in adverse effects on the objectives, including $f_1$ and $f_2$. To investigate the influence of the input ingredients, the amount of roasted barley, which offers a higher SRM value, is increased to $5$ kg (from the original $0.5$ kg) and the optimisation is re-run 30 times for one of the failed products, product 7 (i.e. Sunmaid Stout). The result demonstrates an uptake in using the `topped-up' ingredient (on average $3.12$ kg) to meet the objectives. In summary, the process returns $45 \pm 31$ successful solutions, out of the $85 \pm 20$ non-dominated ones, which is in line with the figures viewed for the other products.

\begin{table}\centering
	\caption{Average distance from the optimal objective values in failed products.  \label{tab:obj-avg-failedProd}}
	\setlength{\tabcolsep}{17pt}
	\begin{tabular}{cccc}
		\toprule 
		\textbf{\#} & \textbf{$f_{1}$} & \textbf{$f_{2}$} & \textbf{$f_{3}$}\tabularnewline
		\midrule 
		4 & 1.054 & 11.032 & 13.181\tabularnewline
		7 & 1.489 & 80.655 & 42.872\tabularnewline
		14 & 1.891 & 117.333 & 29.478\tabularnewline
		\bottomrule 
	\end{tabular}
	
\end{table}

To summarise, in instances where domain experts require more diverse set of solutions, the single objective optimisation method presents a greater promise. On the other hand, the multi-objective optimiser provides a delicate choice of solutions from a particular region within the solution space with inherent proximity in the objective space which allows the exploitation of a promising region of the solution space. Therefore, an overall suitable solutions can be identified through the first method, and then the second can be deployed to fine-tune the favourite solution.

\section{Conclusion}

When aiming for innovation, the high experimental costs associated with the beer brewing process is shown to be efficiently reducible by taking into account the key product characteristics (and reformulating them as objectives) along with the input ingredients. The proposed method automates the quantitative ingredients selection, which is one of the key experimental aspects of brewing. Although we established this primarily with volumes corresponding to low cost production environments, the presented system can be geared towards scalability.
Therefore, the core challenge of generating novel and dynamically changing recipes, based on the product characteristics, is alleviated. This allows the design of high quality beer by commercial venues where quantities and varieties of ingredients are not hard constraints, as well as less equipped and more flexible microbreweries where constraints play a bigger role.
A number of strategies are suggested depending on user preferences in terms of the solution choice, either from a narrow set of solution with delicate differences, or a diverse pool of potential solutions with further discriminability.

More investigation is required to determine the impact of multi-criteria decision-making (MCDM) methods on the existing workflow. An important next step, with a potentially greater impact, is to explore crowd distancing mechanism which may result in building a more comprehensive Pareto front, allowing control on either exploiting a narrow solution space, or covering a diverse set of more radically different solutions. Considering additional and more complex objectives, such as aroma profile, flavour and foam is part of an ongoing research.

\begin{landscape}
\begin{table}
	\caption{Average deviation of individual ingredient uptakes within each independent runs and for each product \label{tab:sol-dis-x1-16}}
	\footnotesize 
	\setlength{\tabcolsep}{3pt}
	\renewcommand{\arraystretch}{1.3}
	\begin{tabular}{ccccccccccccccccc}
		\toprule
		\textbf{\#} & $x_{1}$ & $x_{2}$ & $x_{3}$ & $x_{4}$ & $x_{5}$ & $x_{6}$ & $x_{7}$ & $x_{8}$ & $x_{9}$ & $x_{10}$ & $x_{11}$ & $x_{12}$ & $x_{13}$ & $x_{14}$ & $x_{15}$ & $x_{16}$\tabularnewline
		\midrule
		1 & 3.16E-03 & 1.92E-03 & 3.09E-03 & 1.26E-03 & 1.53E-03 & 1.75E-04 & 2.52E-05 & 5.82E-04 & 1.56E-04 & 6.42E-04 & 1.17E-04 & 1.17E-04 & 4.58E-04 & 2.14E-05 & 7.07E-04 & 7.68E-04\tabularnewline
		2 & 1.29E-03 & 9.52E-04 & 1.26E-03 & 6.65E-04 & 1.26E-03 & 1.80E-04 & 2.82E-04 & 5.52E-03 & 2.17E-04 & 4.62E-05 & 5.11E-03 & 5.11E-03 & 3.21E-04 & 6.43E-04 & 5.78E-05 & 7.14E-04\tabularnewline
		3 & 3.04E-03 & 2.10E-03 & 3.31E-03 & 1.60E-03 & 1.71E-03 & 7.59E-04 & 1.36E-03 & 2.66E-03 & 2.30E-03 & 7.59E-03 & 2.59E-02 & 2.59E-02 & 3.48E-02 & 2.43E-04 & 2.27E-03 & 8.63E-04\tabularnewline
		4 & NA & NA & NA & NA & NA & NA & NA & NA & NA & NA & NA & NA & NA & NA & NA & NA\tabularnewline
		5 & 2.88E-03 & 1.22E-03 & 1.47E-03 & 8.44E-04 & 2.04E-03 & 1.86E-02 & 1.11E-03 & 9.77E-03 & 5.01E-03 & 1.03E-02 & 1.07E-02 & 1.07E-02 & 2.41E-03 & 2.75E-04 & 3.24E-03 & 6.97E-04\tabularnewline
		6 & 1.51E-03 & 8.64E-04 & 1.48E-03 & 6.30E-04 & 2.59E-03 & 3.34E-02 & 9.83E-03 & 2.63E-02 & 1.31E-02 & 5.08E-02 & 1.92E-02 & 1.92E-02 & 4.03E-02 & 7.60E-03 & 6.09E-03 & 8.94E-04\tabularnewline
		7 & NA & NA & NA & NA & NA & NA & NA & NA & NA & NA & NA & NA & NA & NA & NA & NA\tabularnewline
		8 & 1.52E-03 & 8.87E-04 & 1.59E-03 & 6.25E-04 & 1.73E-03 & 9.12E-02 & 1.98E-02 & 1.61E-02 & 8.62E-03 & 7.26E-02 & 3.62E-02 & 3.62E-02 & 4.58E-02 & 3.55E-03 & 1.49E-02 & 9.40E-04\tabularnewline
		9 & 7.71E-04 & 3.47E-04 & 3.77E-04 & 3.02E-04 & 7.64E-04 & 2.93E-02 & 8.67E-04 & 2.25E-02 & 2.32E-03 & 4.34E-02 & 6.05E-03 & 6.05E-03 & 6.25E-02 & 1.79E-04 & 1.14E-02 & 9.51E-04\tabularnewline
		10 & 2.04E-03 & 1.21E-03 & 1.79E-03 & 9.71E-04 & 2.22E-03 & 3.71E-02 & 9.53E-03 & 8.04E-03 & 9.55E-03 & 2.61E-02 & 3.05E-02 & 3.05E-02 & 3.83E-02 & 7.69E-04 & 8.14E-03 & 6.55E-04\tabularnewline
		11 & 1.31E-03 & 6.74E-04 & 9.78E-04 & 4.66E-04 & 1.16E-03 & 3.60E-03 & 3.22E-04 & 6.19E-03 & 9.90E-04 & 2.20E-03 & 1.92E-03 & 1.92E-03 & 2.25E-03 & 1.64E-04 & 1.06E-03 & 5.62E-04\tabularnewline
		12 & 2.97E-03 & 2.05E-03 & 2.64E-03 & 1.47E-03 & 1.69E-03 & 5.12E-04 & 1.59E-04 & 1.52E-03 & 1.47E-04 & 6.46E-04 & 1.06E-03 & 1.06E-03 & 9.57E-04 & 4.60E-05 & 1.12E-03 & 6.33E-04\tabularnewline
		13 & 2.51E-03 & 1.28E-03 & 2.00E-03 & 1.27E-03 & 2.54E-03 & 4.64E-02 & 2.97E-02 & 1.55E-02 & 8.46E-03 & 4.18E-02 & 4.17E-02 & 4.17E-02 & 3.57E-02 & 4.07E-03 & 1.27E-02 & 7.10E-04\tabularnewline
		14 & NA & NA & NA & NA & NA & NA & NA & NA & NA & NA & NA & NA & NA & NA & NA & NA\tabularnewline
		15 & 3.31E-03 & 1.55E-03 & 1.90E-03 & 1.08E-03 & 1.99E-03 & 2.65E-03 & 8.19E-04 & 1.48E-03 & 3.06E-04 & 8.28E-03 & 9.55E-03 & 9.55E-03 & 4.16E-03 & 2.98E-04 & 7.30E-03 & 6.73E-04\tabularnewline
		16 & 1.86E-03 & 1.12E-03 & 1.29E-03 & 8.60E-04 & 1.46E-03 & 3.24E-02 & 2.40E-03 & 1.52E-02 & 6.02E-03 & 1.85E-02 & 1.58E-02 & 1.58E-02 & 3.42E-02 & 1.14E-03 & 1.04E-02 & 1.03E-03\tabularnewline
		17 & 2.91E-03 & 1.14E-03 & 1.83E-03 & 8.60E-04 & 1.32E-03 & 4.39E-03 & 2.97E-03 & 6.81E-03 & 4.09E-03 & 1.52E-03 & 2.63E-03 & 2.63E-03 & 2.96E-03 & 3.67E-04 & 1.21E-03 & 1.02E-03\tabularnewline
		18 & 3.16E-03 & 2.13E-03 & 3.41E-03 & 1.52E-03 & 2.39E-03 & 3.12E-03 & 5.40E-04 & 1.10E-03 & 3.31E-04 & 7.66E-04 & 1.40E-04 & 1.40E-04 & 4.51E-03 & 1.94E-04 & 2.22E-03 & 7.49E-04\tabularnewline
		19 & 9.73E-04 & 7.87E-04 & 1.24E-03 & 5.48E-04 & 1.13E-03 & 1.89E-02 & 1.35E-03 & 1.30E-02 & 1.36E-03 & 7.24E-03 & 1.01E-02 & 1.01E-02 & 2.75E-02 & 1.73E-03 & 2.93E-03 & 9.40E-04\tabularnewline
		20 & 2.84E-03 & 1.61E-03 & 2.00E-03 & 9.92E-04 & 2.81E-03 & 6.25E-02 & 8.08E-04 & 1.72E-02 & 3.23E-03 & 1.47E-02 & 3.18E-03 & 3.18E-03 & 7.17E-02 & 3.50E-04 & 7.52E-03 & 9.20E-04\tabularnewline
		\bottomrule
	\end{tabular}
\end{table}
\end{landscape}

\bibliographystyle{plain}
\bibliography{myRef}

\end{document}